\documentclass{svproc}
\usepackage[backend=bibtex,style=numeric,sorting=none]{biblatex}
\usepackage{graphicx}
\usepackage{geometry}
\usepackage{float}
\usepackage{url}
\usepackage{array}
\newcolumntype{P}[1]{>{\centering\arraybackslash}p{#1}}
\newcolumntype{M}[1]{>{\centering\arraybackslash}m{#1}}

\begin{document}
\mainmatter              
\title{Empirical Study of Symmetrical Reasoning in Conversational Chatbots}
\titlerunning{Symmetrical Reasoning in Chatbots}  
%
\author{Daniela N. Rim$^1$, Heeyoul Choi$^1$$^2$}
%
%
%
\institute{$^1$Handong Global University, Republic of Korea,\\
$^2$\email{heeyoul@gmail.com}
}

\maketitle              

\begin{abstract}
This work explores the capability of conversational chatbots powered by large language models (LLMs), to understand and characterize predicate symmetry, a cognitive linguistic function traditionally believed to be an inherent human trait. Leveraging in-context learning (ICL), a paradigm shift enabling chatbots to learn new tasks from prompts without re-training, we assess the symmetrical reasoning of five chatbots: ChatGPT 4, Huggingface chat AI, Microsoft's Copilot AI, LLaMA through Perplexity, and  Gemini Advanced. Using the Symmetry Inference Sentence (SIS) dataset by Tanchip et al. (2020), we compare chatbot responses against human evaluations to gauge their understanding of predicate symmetry. Experiment results reveal varied performance among chatbots, with some approaching human-like reasoning capabilities. Gemini, for example, reaches a correlation of 0.85 with human scores, while providing a sounding justification for each symmetry evaluation. This study underscores the potential and limitations of LLMs in mirroring complex cognitive processes as symmetrical reasoning.
\end{abstract}
\section{Introduction}
%
Since 2022, the release of AI \textit{conversational chatbots} (also known as \textit{intelligent virtual assistants}) has marked a pivotal moment in human-computer interactions, suggesting a potential paradigm shift in how humans engage with computational systems \cite{mesko2023impact,petroșanu2023tracing,rane2023contribution}. These tools often perform comparably to or exceed human performance in several language understanding tasks \cite{bubeck2023sparks}. The success of the pre-trained large language models (LLMs) that power the chatbots is based on a combination of training data quality, carefully tuned and crafted architectures, optimization techniques, and available computational resources \cite{zhao2023survey,min2023recent}. In particular, the combination of better architectures, optimization, and computational resources allowed these models to increase their capacity in the order of a billion to trillion trainable parameters. This allowed LLMs to train in diverse natural language tasks such as text generation, language translation, diverse text classification, dialogue systems and chatbots, natural language inference (NLI), and several others.

Conversational chatbots were developed based on the multi-task language training of LLMs. Arguably, one of the most famous chatbots is ChatGPT\footnote{\url{https://chat.openai.com/}}, released for the first time in November 30, 2022. Its interface allows users to engage in a wide range of conversational and informational exchanges through prompts. ChatGPT is mainly powered by the LLMs GPT family \cite{brown2020language}, which are in turn based on the Transformer language model \cite{vaswani2017attention}. The high number of parameters in the GPT models and the possibility of prompting allowed a novel paradigm called \textit{in-context learning} (ICL)\cite{brown2020language}, in which models \textit{`make predictions only based on contexts augmented with a few examples'} \cite{dong2022survey}, while previous smaller models required extensive training or fine-tuning in order to learn new tasks. ICL is not restricted to the GPT family, but rather a common characteristic of all LLMs, and is naturally inherited by conversational chatbots. Without the need for re-training or fine-tuning, these chatbots can learn novel tasks from the prompts alone. 

The essence of ICL lies in the ability of LLMs to capture language patterns from the vast training data, and thus being able to adapt and generalize this knowledge to novel tasks \cite{zhao2023survey}. This behavior can be related to the existence of innate mind structures that process, understand and generalize language patterns as theorized in the field of cognitive science. Specifically, the widely supported linguistic theory of Universal Grammar \cite{chomsky2014aspects} defines language as structural representations of linguistic objects with structural laws that govern them. The innate capability of humans to capture such laws is what allows for the naturalization of language behaviors and the ability to apply linguistic knowledge across different contexts. 
Linguistic symmetry is one of such fundamental properties of natural language \cite{gleitman2019emergence} and can be present at word-level, sentence-level and conceptual level \cite{elkin2010symmetry}. 
For instance, the existence of symmetry in language structures is what allows humans to infer \textit{`Mary met John'} from \textit{`John met Mary'} implicitly. 

In this study, we focus on sentence-level predicate symmetry, which aids in systematic relationship inference among entities/contexts \cite{henderson2008symmetry}. In such cases, linguistic objects (usually verbs) relate two or more subjects to an object, serving as transformations that allow symmetrical/asymmetrical relations. For example, the word `marry' immediately allows inferring information: if \textit{A marries B}, then it naturally implies that \textit{B marries A.} However, a word's symmetrical property is not inherent \cite{henderson2008symmetry}, as it also depends on the context being used. The symmetry inference of the word \textit{kissed} is not the same in \textit{`John and Mary kissed'} as it is in \textit{`John kissed Mary.'} For this reason, the symmetrical property of a word cannot be deterministically assigned and requires human cognition for its categorization. 
Given that conversational chatbots are viewed as analogous to humans, we aim to investigate their ability to grasp predicate symmetry.

In this work, we take advantage of ICL and test the capability of predicate symmetry understanding and characterization from 5 conversational chatbots: ChatGPT 4, Huggingface chat AI, Microsoft's Copilot AI, LLaMA through Perplexity, and  Gemini Advanced. Our goal is to empirically observe symmetrical reasoning capabilities from answering ``does \textit{`A [related to] B'} imply \textit{`B [related to] A}?" To achieve this goal, we used the symmetry inference sentence (SIS) dataset introduced by Tanchip et al. (2020) \cite{tanchip-etal-2020-inferring}. We contrast the conversational chatbots symmetrical evaluation with human evaluation scores. The results show that while some AI assistants perform poorly in comparison to human evaluations, others have a high correlation and perform at a competitive level. To our knowledge, this work is the first one to test the symmetrical reasoning capability in LLMs.

The paper is organized as follows: Section \ref{relwork} details related works on symmetry in linguistic theory and in the context of large language models. Section \ref{method} describes our methodology, including the dataset used for the experiments and the prompts designed for the conversational chatbots. Finally, Section \ref{experiments} details our experiments and presents the results.

\section{Related Work} \label{relwork}

\subsection{Linguistic Symmetry}
We are interested in the concept of symmetry in language. In general, the concept of symmetry plays a key role in domains like nature (e.g. physics), mathematics, and visual perception. For example, symmetry in physics is deeply linked to conservation laws, allowing the behavioral prediction of systems without complex calculations. The simplification of calculation in mathematical demonstrations is also an advantage in the mathematical field deriving from symmetrical properties. It also extends to the realm of language, where it influences the way we use, adapt, and generalize abstract concepts \cite{weyl2015symmetry}. Linguistic symmetry, much like symmetry in mathematics, can be defined as the invariance of meaning under specific transformations \cite{henderson2008symmetry}. Nonetheless, `invariance of meaning' is a challenging criterion to define and quantify in theoretical linguistics, unlike mathematical transformations. There is, however, evidence that symmetrical transformations exist and obey linguistic structural laws \cite{dimitriadis2008event, siloni2012reciprocal}.

In this work, we investigate sentence-level symmetry, i.e., within syntactic structures. In particular, linguistic alternations showcase symmetry when the core meaning of a sentence is maintained despite slight changes in word order or phrasing. One example is diathesis alternations, where a verb can be used in various syntactic frames while retaining the overall predicate-argument relation (e.g., `\textit{he read a book}' vs `\textit{the book was read by him}'). A work by Lapata (1999) \cite{lapata1999acquiring} has shown that these alternations relate to linguistic intuitions and lexical generalizations such as establishing systematic relationships between verb meaning and argument realization.

\subsection{Linguistic Symmetry in Language models}

Our work is related to the study by Tanchip et al. \cite{tanchip-etal-2020-inferring} on the capability of language models to rate the symmetrical property of certain predicates. They claim that a model that can distinguish symmetry relations is able to systematically generalize better in the case of Natural Language Inference tasks, which often miss the nuanced contributions of certain word classes or semantic regularities \cite{goodwin2020probing}. The authors hypothesize the importance of contextualized information as an aid to identify symmetry, and thus use the pre-trained language model BERT \cite{devlin2018bert} to include such information in the symmetry scoring.

In order to evaluate their hypothesis, the authors provided a novel symmetry inference sentence dataset
(SIS) containing sentence pairs and a symmetry score given by human evaluators. Their best performing model was a hybrid combination of a regression layer (to capture linguistic features) and a fine-tuned BERT encoder (to add contextual knowledge), concluding that while contextualized language models are effective in applying a context-based approach to symmetry inference, they lack in capturing essential linguistic features related to symmetry. 

Our work can be considered as an extension of Tanchip et al. \cite{tanchip-etal-2020-inferring} which was published before the existence of conversational chatbots. We use the provided dataset with human scores to explore the symmetrical reasoning capabilities through ICL in the chatbots. We empirically evaluate if the conversational chatbots perform similarly to humans in this specific task without the need of fine-tuning or re-training.

\section{Methodology} \label{method}

In this section, we describe the prompt design to input to the conversational chatbots, as well as the conversational chatbots that were used.

\begin{table}[H]
\caption{Two verb examples (of 40) occurring in three (out of 10) different contexts drawn from the SIS dataset. Each sentence (a) is given alongside an alternated version (b).} 
\begin{center}
    \begin{tabular}{lp{3.7cm}p{3.7cm}p{3.7cm}} \hline
\multicolumn{1}{c}{\textbf{Verb}} & \multicolumn{1}{c}{\textbf{Example 1}}     & \multicolumn{1}{c}{\textbf{Example 2}}            & \multicolumn{1}{c}{\textbf{Example 3}}  \\ \hline
\textit{mix}                       & \begin{tabular}[c]{@{}p{3.7cm}@{}}(a) The goths and punks mix with each other quite well.\\ (b) The punks and goths mix with each other quite well.\end{tabular}                   & \begin{tabular}[c]{@{}p{3.7cm}@{}}(a) She mixed her facades and garden ornament knick knacks.\\ (b) She mixed her garden ornament knick knacks and facades.\end{tabular}               & \begin{tabular}[c]{@{}p{3.7cm}@{}}(a) Oil and water don't mix.\\ (b) Water and oil don't mix.\end{tabular}                                   \\[2pt] \hline
\textit{love}                      & \begin{tabular}[c]{@{}p{3.7cm}@{}}(a) Siyabonga will love it but it won't love him.\\ (b) It will love Siyabonga but it won't love him.\end{tabular}                               & \begin{tabular}[c]{@{}p{3.7cm}@{}}(a) How can Adam and Eve love as Christ loved?\\ (b) How can Eve and Adam love as Christ loved?\end{tabular}                                         & \begin{tabular}[c]{@{}p{3.7cm}@{}}(a) Mothers and priests love unconditionally.\\ (b) Priests and mothers love unconditionally.\end{tabular} \\ [2pt]
\hline
\end{tabular}
\end{center}
\label{table:examples}
\end{table}

\subsection{Prompt Design}\label{prompts}

The SIS dataset \cite{tanchip-etal-2020-inferring} consists of 400 sentences, with 10 different examples of 40 seed verbs identified by linguists as symmetrical and asymmetrical from the English Web 2015 Corpus \cite{jakubivcek2013tenten}. Each sentence has at least two entities and a verb that relates them so that it can be syntactically alternated (words are rearranged in a linguistic-sounding way.) A few examples of this dataset are presented in Table \ref{table:examples}. This raw dataset can be fed to different models to obtain several benchmark scores. For instance, word feature-based scores (from linguistic theory), contextual scores resulting from averaging token representations using the pre-trained models Word2Vec \cite{mikolov2013efficient} and GloVe \cite{pennington2014glove}, raw and fine-tuned BERT, and any other models. 

In our work, instead of using this raw dataset version of SIS, the conversational chatbots are prompted with an adapted version of the SIS online survey given to human evaluators (see \cite{tanchip-etal-2020-inferring} for details of the online survey.) That is, the conversational chatbots are given the same general direction as follows:

\vspace{5pt}

\noindent\fbox{%
    \parbox{0.98\textwidth}{%
        \textbf{Prompt 1}\\
        \textit{A sentence is symmetrical if all participants are simultaneously on the giving and receiving end of the action described. If you switch the position of the participants, the overall meaning of the sentence won’t change. In this task, you will be given a pair of sentences. The first describes at least two participants and describes their relationship. The second sentence conveys the same information as the first, except the positions of the participants in the sentence are switched. Your task is to rate how alike in meaning the given two sentences are from a scale of 1-5, where 1 means the sentences do mean the same and 5 means do not mean the same}
    }%
}

\vspace{5pt}

After this prompt was given, all five models understood the task and expressed their readiness to answer the prompts. Next, an example was provided to the chatbots as follows:

\vspace{5pt}

\noindent\fbox{%
    \parbox{0.98\textwidth}{%
    \textbf{Prompt 2}\\
    \textit{Given the following pair of sentences:
        \begin{itemize}
            \item[(a)] \textit{Sample from SIS dataset}
            \item[(b)] \textit{Alternate from (b)}
        \end{itemize}
        Rate how alike in meaning the given two sentences are from a scale of 1-5, where 1 means the sentences do mean the same, and 5 means the sentences do not mean the same.}
    }%
}

\vspace{5pt}

All the chatbots would reply to \textbf{Prompt 2} with a rating and corresponding verbal justification, to which a third prompt would be given:

\vspace{10pt}

\noindent\fbox{%
    \parbox{0.98\textwidth}{%
    \textbf{Prompt 3}\\
    Please return just the numerical score without an explanation of reasoning.
    }%
}

\vspace{5pt}

Finally, the conversational chatbots would output the score alone. From that prompt onwards, the 400 survey questions following the format of \textbf{Prompt 2} would be fed recursively and the answers collected.

\subsection{Conversational Chatbots}

The conversational chatbots used for this symmetry evaluation are ChatGPT 4\footnote{\url{https://chat.openai.com/}}, HuggingChat\footnote{\url{https://huggingface.co/chat/}}, Copilot\footnote{\url{https://copilot.microsoft.com/}}, Perplexity\footnote{\url{https://www.perplexity.ai/}}, and Gemini Advanced\footnote{\url{https://gemini.google.com/}}. In Table \ref{table:LLMs} we detailed the characteristics of the LLMs that powered each AI.

\begin{table}[H]
\caption{Detailed information of the used conversational chatbots. The \textit{Latest Update} are reported on the last day the experiments from this paper were conducted (March 7th, 2024.) The LLM column denotes the LLM model that the conversational chatbot is mainly based on. The number of parameters of each LLM is added (reported or estimated.)}
\begin{center}
    \begin{tabular}{lccc}
\hline
\multicolumn{1}{c}{\textbf{\begin{tabular}[c]{@{}c@{}}Conversational\\ Chatbot\end{tabular}}} & \textbf{\begin{tabular}[c]{@{}c@{}}Latest \\ update\end{tabular}} & \textbf{LLM} & \textbf{\begin{tabular}[c]{@{}c@{}}Number of \\ parameters (B)\end{tabular}} \\ \hline
\textbf{ChatGPT 4} & Feb. 13, 2024 & GPT 4 \cite{achiam2023gpt} &  \textless 175       \\
\textbf{HuggingChat}     & Oct. 4, 2023       & Mixtral 8x7B \cite{jiang2024mixtral} & 47   \\
\textbf{Copilot}  & March 7, 2024    & GPT 4 \cite{achiam2023gpt}     & \textless 175    \\
\textbf{LLaMA with Perplexity}    & -    & LLaMA 2 \cite{touvron2023llama}      & 65    \\
\textbf{Gemini Advanced}   & March 3, 2024   & Gemini \cite{team2023gemini}  & undisclosed  \\                        \hline                                  
\end{tabular}
\end{center}
\label{table:LLMs}
\end{table}

We chose these conversational chatbots as they are supported by renowned companies (such as Google, Microsoft, Open AI, and others) and are available to the general public. Other chatbots such as \textit{xAI Grok} \footnote{\url{https://grok.x.ai/}, visited until March 6th, 2024.} had limited access to certain users (in this case, X Premium+ users.) Other conversational chatbots of lesser-known backgrounds were left out due to poor performances (showing poor understanding of the prompts presented in Section \ref{prompts}.)

\section{Experiments} \label{experiments}
The conversational chatbots ChatGPT 4 (from now on ChatGPT), Gemini Advanced (from now on Gemini), HuggingChat, Copilot, and Perplexity were equally given the task explanation of \textbf{Prompt 1} (Section \ref{prompts}) plus the 400 prompts containing the symmetrical sentences to be evaluated (\textbf{Prompt 2} followed by \textbf{Prompt 3}, and then successively \textbf{Prompt 2} with the SIS dataset). The answers to \textbf{Prompt 2} (numerical scores from each data pair) are collected for each chatbot.

The resulting scores are compared with the SIS evaluation scores given by \cite{tanchip-etal-2020-inferring}: human evaluation, feature prediction, Word2Vec, GloVe, (raw and fine-tuned) BERT, and hybrid. The human evaluation was obtained by averaging the scores from 1 to 5 given by seven evaluators for the SIS 400 questions, the same with the prompts 1 and 2 given to the chatbots. The feature prediction scores were retrieved from pre-defined linguistic features (see \cite{tanchip-etal-2020-inferring} for more details.) Word2Vec and GloVe scores were retrieved by averaging the token embeddings of each sentence case in the dataset. The symmetry scores given by the BERT were retrieved after forwarding the sentences through the model and obtaining the representation of the special [SEP] token \cite{devlin2018bert}. The BERT was fine-tuned using the symmetrical feature score given by linguistics. Finally, `hybrid' is the combination of pre-training BERT as an encoder, with a regression model on top to classify the symmetry score (proposed method by \cite{tanchip-etal-2020-inferring}.) 

Since conversational chatbots are stochastic, we selected the best-performing chatbot from the prompting trials previously explained and repeated the questionnaire seven times (like the number of human evaluators.) We observed if the chatbot's answers had high variation between them when confronted with the same prompt, possibly hinting at randomness and chance in their decision. The results will be reported in the next section.

\subsection{Results}

Table \ref{table:prompt1_ex} shows the five conversational chatbots' answers to \textbf{Prompt 1}. ChatGPT and Copilot answered briefly, expressing their understanding of the task and expecting the following prompts with the scoring question. The similarity of their answer structure could be attributed to the fact that both chatbots have GPT4 as their core model as shown in Table \ref{table:LLMs}. HuggingChat and Gemini also answered similarly, expressing not only their understanding but also providing an example of the expected questions. Gemini even generated a detailed explanation of the reasoning behind each score. In comparison to these four, Perplexity showed a poor understanding of the prompt, choosing to expand on the concepts mentioned in \textbf{Prompt 1}, such as symmetry in sentences and rating scales. Overall, Gemini provided more compelling evidence of a clear understanding of the upcoming task.

\begin{table}[H]
\caption{Output of the conversational chatbots when confronted with \textbf{Prompt 1} in Section \ref{prompts}.}
\begin{tabular}{M{2.3cm}|p{10.5cm}}
\hline
\multicolumn{1}{c|}{\textbf{\begin{tabular}[c]{@{}M{2.3cm}@{}}Conversational\\ Chatbot\end{tabular}}} &
  \textbf{Answer to Prompt 1} \\ \hline
\textbf{ChatGPT} &
  I understand your task. Please provide the pair of sentences you'd like me to evaluate for similarity in meaning based on the position of the participants. \\ \hline
\textbf{HuggingChat} &
  \begin{tabular}[c]{@{}p{10.5cm}@{}}Sure! I'll do my best to accurately rate the similarity in meaning between each pair of sentences based on your instructions.\\ Here are some examples:\\ Sentence 1: "John gave Sarah a gift."\\ Sentence 2: "Sarah received a gift from John." (Rating: 1)\\ These two sentences describe the same event but with reversed roles. They have an identical underlying meaning.\end{tabular} \\ \hline
\textbf{Perplexity} &
  \begin{tabular}[c]{@{}p{10.5cm}@{}}The concept of symmetry in sentences refers to sentences where participants are both giving and receiving the action described, and switching their positions does not change the overall meaning. \\ In the context of rating scales, Likert scales are commonly used, which aim to be symmetrical with an equal number of positive and negative responses. \\ These scales typically range from 1 to 5, allowing respondents to rate severity or magnitude, with 1 being the lowest and 5 being the highest rating. The Likert scale is a bipolar scaling method where respondents indicate their level of agreement or disagreement on a symmetric agree-disagree scale for a series of statements. It is important to keep Likert scales odd to include a midpoint and prevent respondents from falling into a rhythm, thus avoiding biased responses.\\ In summary, when using rating scales like Likert scales that range from 1 to 5, maintaining symmetry in the scale design by ensuring an equal  balance of positive and negative responses is crucial for accurate data collection and analysis\end{tabular} \\ \hline
\textbf{Copilot} &
  Certainly! Please provide the pair of sentences, and I’ll evaluate their similarity in meaning. (smileyface) \\ \hline
\textbf{Gemini} &
  \begin{tabular}[c]{@{}p{10.5cm}@{}}Absolutely!  Here's how I understand the task and the rating system:\\ Task\\ I'll receive pairs of sentences. The goal is to determine if the sentences are symmetrical, meaning they have the same meaning even if the order of the participants is switched.\\ Rating System\\ 1 - Sentences have the exact same meaning.\\ 2 - Sentences have mostly the same meaning, with potential minor differences in nuance.\\ 3 - Sentences have similarities in meaning, but there are noticeable differences.\\ 4 - Sentences share a core concept, but have significant meaning differences.\\ 5 - Sentences have completely different meanings.\\ Example\\ Sentence 1: Sarah hugs Ben.\\ Sentence 2: Ben hugs Sarah.\\ Rating: 1 (The meaning is exactly the same despite the word order change)\\ Let's get started! Please provide the first pair of sentences.\end{tabular} \\
  \hline
\end{tabular}
\label{table:prompt1_ex}
\end{table}

We show the correlation of each method and model in \cite{tanchip-etal-2020-inferring} and conversational chatbot in the SIS dataset with the average human score in Table \ref{table:cor_scores}. HuggingChat and Gemini scores show evidence of ICL, that is, with just prompting, these chatbots performed comparatively to fine-tuning BERT in a natural language inference task. Please note that the high score of the hybrid method (\cite{tanchip-etal-2020-inferring}) is due to fine-tuning BERT and training an extra layer to extract features related to symmetry. In contrast, no additional symmetrical information was given to the conversational chatbots to aid them in their judgment.

\begin{table}[H]
\caption{Correlation scores of each of the methods/models/chatbots with the averaged human scores. `\textit{CC. Avg.} is the correlation of the five conversational chatbots averaged scores. The `$*$' denotes the scores given by \cite{tanchip-etal-2020-inferring}.}
\centering
\begin{tabular}{cccccc}
\hline
\textbf{BERT*}    & \textbf{BERT-finet.*} & \textbf{Ft. Pred.*} & \textbf{Word2Vec*}   & \textbf{GloVe*}  & \textbf{Hybrid*}           \\ \hline
0.488            & 0.790                & 0.593              & 0.319               & 0.325           & 0.903             \\
\hline
\textbf{ChatGPT} & \textbf{HuggingChat} & \textbf{Copilot}   & \textbf{Perplexity} & \textbf{Gemini} & \textbf{CC. Avg.} \\ \hline
0.640            & 0.717                & 0.441              & 0.466               & 0.802           & 0.811      \\
\hline
\end{tabular}
\label{table:cor_scores}
\end{table}

Gemini was the best-performing model in terms of correlation with human evaluation. As explained in the introduction of this section, we evaluated Gemini seven times and compared the averaged scores with the ones given by the seven human evaluators. We show the results in Table \ref{table:cor_mv}. While we do not have access to the individual answers of each human evaluator, we have the vote count for each score (1-5) given each SIS sentence pair. This made it possible to calculate the majority vote (M.V.), i.e., scoring the sentence pair symmetry with the most elected integer out of the five. We used the same metric for the five conversational chatbots (M.V. CC). Additionally, we provide the majority vote of the seven Gemini trials (M.V. Gemini.) As observed in Table \ref{table:cor_mv}, the majority vote correlation of Gemini's trials with human scores reflects the Gemini's potential for generating symmetrical reasoning and judgment. 

\begin{table}[H]
\centering
\caption{Correlation with human scores of the averaged Gemini's 7 trials (Avg. Gemini), the averaged individual correlations of each Gemini trial (Avg. Cor. Gemini), human Majority Vote (M.V. human), Conversational Chatbot majority vote (M.V. CC), the seven Gemini trials majority vote (M.V. Gemini.)}
\begin{tabular}{cc|ccc}
\hline
\textbf{Avg. Gemini} & \textbf{Avg. Cor. Gemini} & \textbf{M.V. human} & \textbf{M.V. CC} & \textbf{M.V. Gemini} \\ \hline
0.848                & 0.772                     & 0.932                & 0.754             & 0.857        \\
\hline
\end{tabular}
 \label{table:cor_mv}
\end{table}

\begin{figure}[H]
  \centering
  \begin{minipage}[b]{0.49\textwidth}
    \includegraphics[width=\textwidth]{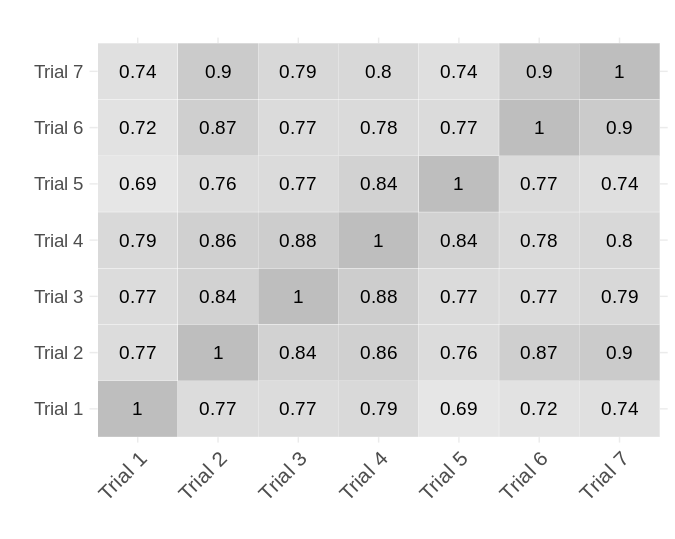}
    \caption{Correlation of the seven Gemini trials, using the same 400 prompts.} \label{figure:corr7}
  \end{minipage}
  \hfill
  \begin{minipage}[b]{0.49\textwidth}
    \includegraphics[width=\textwidth]{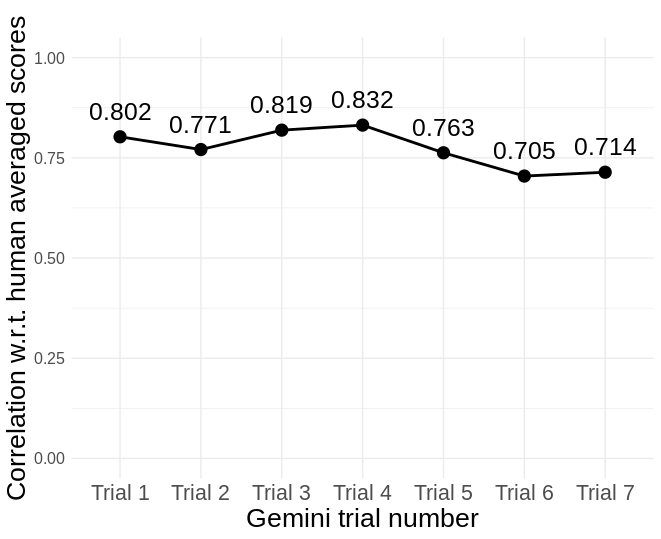}
    \caption{Correlation of the seven Gemini trials with the averaged human scores.} \label{figure:7sym}
  \end{minipage}
\end{figure}

We computed the correlation between the seven trials to determine if Gemini was stable in its answers as presented in Figure \ref{figure:corr7}. As shown, there is a strong correlation within Gemini's answers, which hints that the model made the scoring intentionally. 

Finally, we compared the correlation of each Gemini trial with the humans' averaged scores (Figure \ref{figure:7sym}.) The correlation falls within a threshold of 0.7 and 0.84, showing consistency in Gemini's judgment.

\subsubsection{Qualitative Examples}

Gemini would always provide explanations on its rating scores if not prompted otherwise. We show two output examples to illustrate the evaluation process of Gemini when providing the scores, which enforces the intentionality of each score. The first example is for a sentence pair case judged as a 1 for most human voters and received an average of 1.43.

\vspace{5pt}

\noindent\fbox{%
    \parbox{0.98\textwidth}{%
        \textbf{Prompt}\\
        \textit{Given the following pair of sentences:
        \begin{itemize}
            \item[(a)] He married Kristie and has three children, Afiya, Eden and Titus. 
            \item[(b)] Kristie married him and has three children, Afiya, Eden and Titus. 
        \end{itemize}
        Rate how alike in meaning the given two sentences are from a scale of 1-5, where 1 means the sentences do mean the same, and 5 means the sentences do not mean the same. }\\
        
        \textbf{Gemini Advanced answer:}\\
         Here's how I'd rate the similarity of the sentences and my reasoning:
         
         \textbf{Rating:} 1 (The sentences mean the same)  
         \begin{itemize}
             \item \textbf{Core Meaning}: Both sentences describe the same central event: a marriage between two people (implied to be a man and Kristie) and the fact that they have three children together.
             \item \textbf{Perspective Shift}: The primary difference is the perspective. Sentence (a) focuses on ``he" as the subject, while sentence (b) centers on Kristie.
             \item \textbf{Synonym Use:} The words ``married" and "has" convey the same core relationship and situation.
         \end{itemize}
    }%
}

\vspace{5pt}
We also provide an example of a dissimilar score of 4 given by Gemini. The model not only provides an analysis of the alternation in the sentences, but it also provides a philosophical interpretation of the content that would justify a certain degree of symmetry (and thus a 4 instead of a 5) given the sentence's context. On the other hand, the majority of human voters chose this pair to score a 5 in dissimilarity, with an average of 4.285.
\vspace{5pt}

\noindent\fbox{%
    \parbox{0.98\textwidth}{%
        \textbf{Prompt}\\
        \textit{Given the following pair of sentences:
        \begin{itemize}
            \item[(a)] Things might follow natural laws when given the option.
            \item[(b)] Natural laws might follow things when given the option.
        \end{itemize}
        Rate how alike in meaning the given two sentences are from a scale of 1-5, where 1 means the sentences do mean the same, and 5 means the sentences do not mean the same. }\\
        
        \textbf{Gemini Advanced answer:}\\
         I'd rate the similarity of these sentences as follows:         
         \textbf{Rating:} 4 (The sentences have somewhat different meanings)

        Here's my reasoning:
         \begin{itemize}
             \item \textbf{Inverted Relationship}: The sentences reverse the traditional subject-object relationship between ``things" and ``natural laws". Sentence (a): Implies that ``things" have agency and might choose to adhere to natural laws. Sentence (b): Suggests that natural laws themselves could adapt or change in response to ``things".
             \item \textbf{Philosophical Implications}: This inversion touches on philosophical debates about the nature of reality: Does the universe operate on fixed principles that things must follow? Or, are these "laws" more like guidelines that emerge from how things interact?
             \item \textbf{Ambiguity:} The phrase ``when given the option" adds ambiguity to both sentences. It's unclear whether this option is a genuine choice or merely a condition.
         \end{itemize}
    }%
}

\vspace{5pt}

\section{Conclusion}
This study highlights symmetry as one essential feature in human language cognition and its potential presence in advanced conversational chatbots. Our empirical investigation using ICL on the SIS dataset revealed that specific chatbots demonstrate an aptitude for symmetrical reasoning. This finding underscores the increasing sophistication of large language models (LLMs) and their ability to mirror aspects of human cognition.

The tested conversational chatbots varied in performance and understanding of symmetry. This suggests that while LLMs show promise, their grasp of linguistic nuances may still be uneven. Notably, Gemini and HuggingChat exhibited ICL capabilities comparable to fine-tuned language models like BERT, even without explicit training on symmetry features. Gemini, in particular, displayed a strong correlation with human evaluators, demonstrating its aptitude for symmetrical reasoning and judgment. This finding calls for further study on the mechanisms by which LLMs may implicitly acquire and process complex linguistic patterns.

Future work should investigate the implications of stochasticity observed in chatbot responses, understanding the degree of randomness and its potential impact on reliability is crucial. Furthermore, a wider exploration of other human linguistic features and their mirroring within LLMs could deepen the understanding of the boundaries and overlap between human and machine-based language processing.

\section{Acknowledgements}
This research was supported by Basic Science Research Program through the National Research Foundation of Korea funded by the Ministry of Education (NRF-2022R1A2C1012633).

\printbibliography 

%

\end{document}